\pdfoutput=1
\documentclass[11pt]{article}

\usepackage[]{acl}

\usepackage{times}
\usepackage{latexsym}

\usepackage[T1]{fontenc}
\usepackage[utf8]{inputenc}

\usepackage{microtype}
\usepackage{graphicx}
\usepackage{amsfonts}
\usepackage{booktabs}
\usepackage{multirow}
\usepackage{amssymb}
\usepackage{amsmath}
\usepackage[utf8]{inputenc}
\usepackage{cleveref}
\crefname{section}{§}{§§}
\usepackage{subfigure}
\usepackage{array}
\usepackage{color}

\usepackage{hyperref}

\title{JointLK: Joint Reasoning with Language Models and Knowledge Graphs for Commonsense Question Answering}

\author{Yueqing Sun, Qi Shi, Le Qi, Yu Zhang\thanks{\ \ Corresponding author.}\\
Research Center for Social Computing and Information Retrieval\\
Harbin Institute of Technology, Harbin, China\\
        \texttt{\{yqsun,qshi,lqi,zhangyu\}@ir.hit.edu.cn}\\}
\begin{document}
\maketitle
\begin{abstract}
Existing KG-augmented models for commonsense question answering primarily focus on designing elaborate Graph Neural Networks (GNNs) to model knowledge graphs (KGs). However, they ignore (i) the effectively fusing and reasoning over question context representations and the KG representations, and (ii) automatically selecting relevant nodes from the noisy KGs during reasoning. In this paper, we propose a novel model, JointLK, which solves the above limitations through the joint reasoning of LM and GNN and the dynamic KGs pruning mechanism.
Specifically, JointLK performs joint reasoning between LM and GNN through a novel dense bidirectional attention module, in which each question token attends on KG nodes and each KG node attends on question tokens, and the two modal representations fuse and update mutually by multi-step interactions. Then, the dynamic pruning module uses the attention weights generated by joint reasoning to prune irrelevant KG nodes recursively. We evaluate JointLK on the CommonsenseQA and OpenBookQA datasets, and demonstrate its improvements to the existing LM and LM+KG models, as well as its capability to perform interpretable reasoning\footnote{Our code is available at: \href{https://github.com/Yueqing-Sun/JointLK}{https://github.com/Yueqing-Sun/JointLK}}.

% Our results on the CommonsenseQA and OpenBookQA datasets demonstrate that our modal fusion and knowledge pruning methods can make better use of relevant knowledge for reasoning.

\end{abstract}

\begin{figure}[t]
\centering
\includegraphics[width=1\columnwidth]{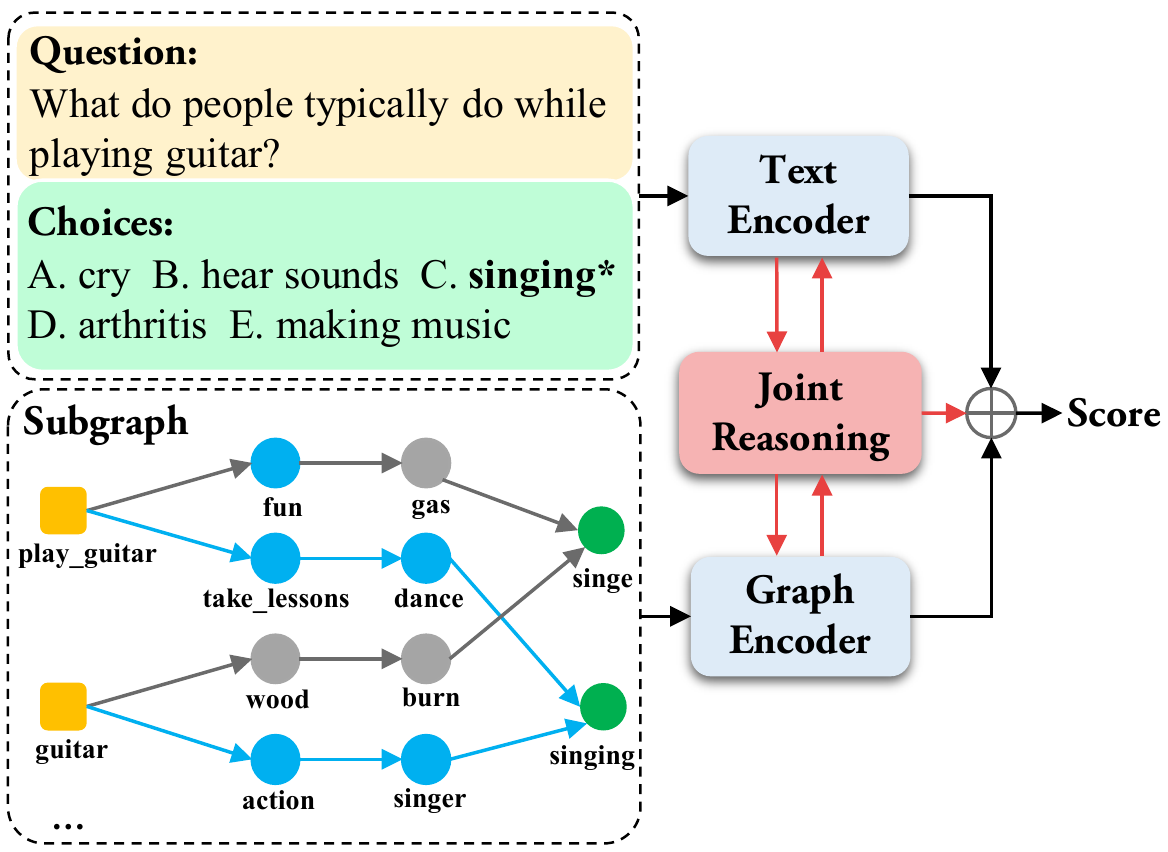} % Reduce the figure size so that it is slightly narrower than the column.
\caption{Our knowledge-augmented joint reasoning model framework with an example from CommonsenseQA. The subgraph is retrieved from ConceptNet.}
\label{fig1}
\end{figure}

\section{Introduction}
Commonsense question answering (CSQA) requires systems to acquire different types of commonsense knowledge and reasoning skills, which is normal for humans, but challenging for machines \cite{talmor-etal-2019-commonsenseqa}. Recently, large pre-trained language models (LMs) have achieved remarkable success in many QA tasks and appear to use implicit (factual) knowledge encoded in their model parameters during fine-tuning \cite{liu2019roberta, JMLR:v21:20-074}. Nevertheless, commonsense knowledge is self-evident to humans and is rarely expressed clearly in natural language  \cite{gunning2018machine}, which makes it difficult for LMs to learn commonsense knowledge from the pre-training text corpus alone.

An extensive research path is to elaborately design graph neural networks (GNNs) \cite{scarselli2008graph} to perform reasoning over explicit structural common sense knowledge from external knowledge bases \cite{10.1145/2629489,speer2017conceptnet}. Related methods usually follow a retrieval-and-modeling paradigm. First, the knowledge subgraphs or paths related to a given question are retrieved by string matching or semantic similarity; such retrieved structured information indicates the relation between concepts or implies the process of multi-hop reasoning. Second, the retrieved subgraphs are modeled by a well-designed graph neural network module \cite{lin-etal-2019-kagnet,feng-etal-2020-scalable,yasunaga-etal-2021-qa} to perform reasoning
over knowledge graphs.

However, these approaches have two main issues. First, \textbf{the retrieved knowledge subgraph contains many noisy nodes}. Whether through simple string matching or semantic matching, in order to retrieve sufficient relevant knowledge, noise knowledge graph nodes will inevitably be included \cite{lin-etal-2019-kagnet,yasunaga-etal-2021-qa}. Especially with the increase of hop count, the number of irrelevant nodes will expand dramatically, raising the burden of the model. As the example in Figure \ref{fig1}, some graph nodes such as ``\textit{wood}", ``\textit{burn}", and ``\textit{gas}", although related to some entities in the questions and choice, can mislead the global understanding of the question. Second, \textbf{there are limited interactions between language representation and knowledge graph representation}. Specifically, existing LM+KG methods \cite{lin-etal-2019-kagnet, feng-etal-2020-scalable} model question context and knowledge subgraphs in isolation by LMs and GNNs, and perform only one interaction in a shallow manner to fuse their representations at the output for prediction.
We argue that the limited interaction between the two modalities is the main bottleneck that may prevent the model from understanding the complex question-knowledge relations necessary to answer the question correctly.

\begin{figure*}[t]
\centering
\includegraphics[width=0.9\textwidth]{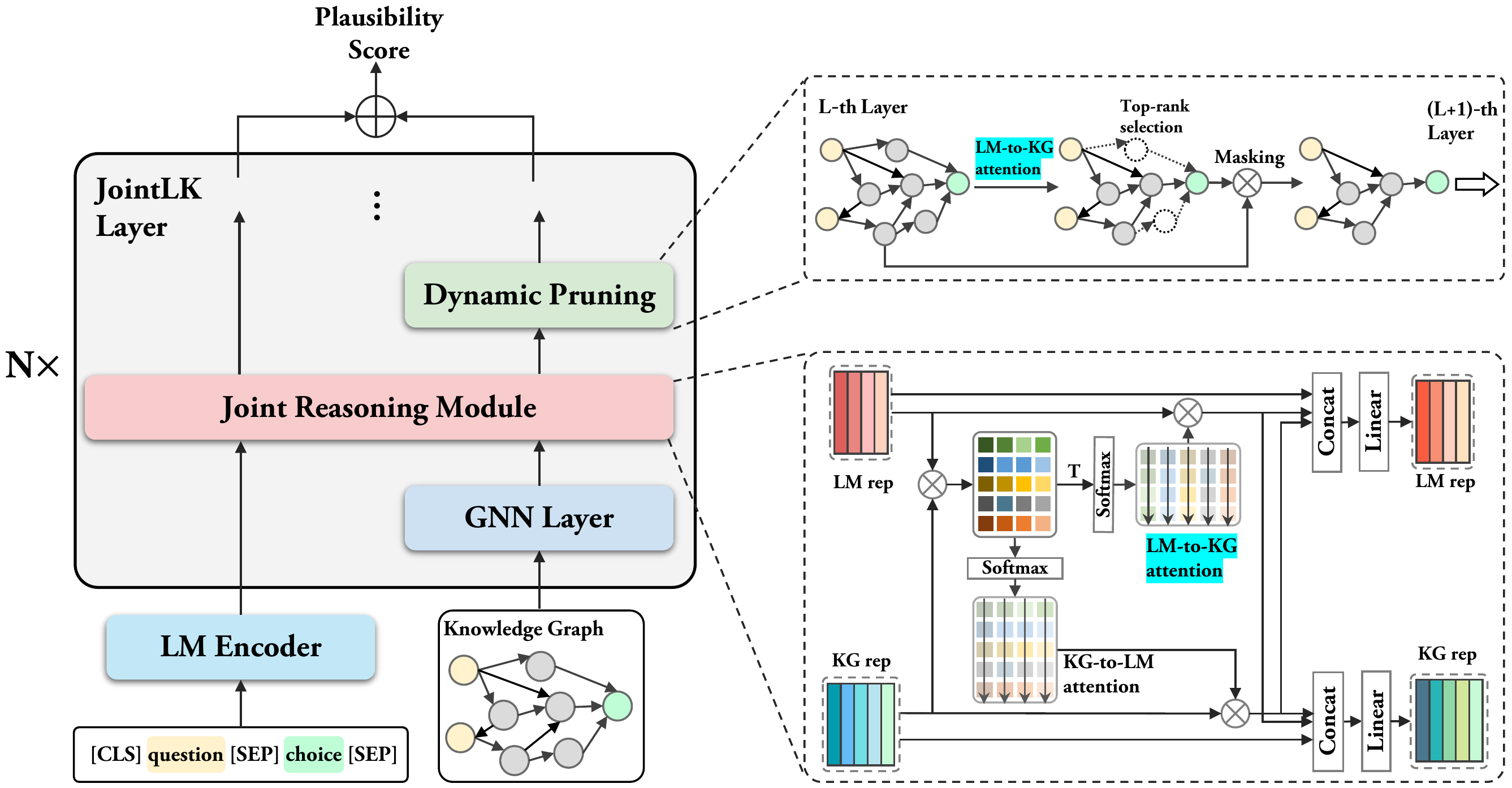} 
\caption{Overall architecture of our proposed JointLK model, which takes a query (question + choice) and a retrieved knowledge subgraph as input, and outputs a scalar that represents the plausibility score of this query. JointLK mainly consists of four modules the Query Encoder, the Graph Layer, the Joint Reasoning Module and the Dynamic Pruning Module, of which the latter three form a stack of N identical layers. }
\label{model}
\end{figure*}

Based on the above consideration, we propose JointLK, a model that performs the fine-grained modal fusion and multi-layer joint reasoning between the language model and the knowledge graph (see Figure \ref{model}).
Specifically, given a question and retrieved subgraphs, JointLK first obtain the representations of the two modalities by using an LM encoder and a GNN encoder respectively. Then we design a joint reasoning module to generate fine-grained bidirectional attention maps between each question token and each KG node to fuse the information from each modality to the other. Guided by the attention generated in the interaction process, the dynamic pruning module deletes irrelevant nodes to make the model reason along the correct knowledge path. Multiple JointLK layers are stacked to form a hierarchy that supports multi-step interactions and recursive pruning. In summary, our contributions are three-fold:
\begin{itemize}
\item We propose JointLK, a novel model that supports multi-step joint reasoning between LM and KG. It uses dense bidirectional attention to simultaneously update query-aware knowledge graph representation and knowledge-aware query representation, bridging the gap between the two information modalities.
\item We design a dynamic graph pruning module that recursively removes irrelevant graph nodes at each JointLK layer to ensure that the model reasons correctly with complete and appropriate evidence.
\item Experimental results show that JointLK is superior to current LM+KG methods, and the refined evidence is interpretable. Furthermore, through the multi-layer fusion of these two modalities, JointLK exhibits strong performance over previous state-of-the-art LM+KG methods in performing complex reasoning, such as solving questions with negation and complex questions with more entities.
\end{itemize}

\section{Related Work}
Commonsense question answering is challenging because the required commonsense knowledge is rarely given in the context of questions and answer choices or encoded in the parameters of pre-trained LMs. Therefore, many works obtain the required knowledge from external sources (e.g., KGs, corpus) to augment CSQA models. 
Due to the heterogeneity between structured knowledge and unstructured text questions, there are currently two main research methods. Some works \cite{lv2020graph,bian2021benchmarking,xu-etal-2021-fusing} unify the two modalities during model input, such as transforming structured knowledge into plain text through templates or transforming question context into structured graphs. However, the original structural/textual information will inevitably be lost during the conversion process. Other works \cite{lin-etal-2019-kagnet,feng-etal-2020-scalable,yan-etal-2021-learning} use LM and GNN to model the two modalities separately, and perform shallow interactions in the latter model stage, such as attentive pooling or simple concatenation of the two modal representations. Although this method can retain the original information of question context and KGs, the limited interaction will affect the flow of information between the two modalities, so we mainly improve on this point.

Recently, QA-GNN \cite{yasunaga-etal-2021-qa} explicitly views the QA context as an additional node, connects it and KG to form a joint graph, and mutually updates their representations through graph-based message passing. However, it pools the representation of the question context into a single node, which limits the updating of the text representation and fine-grained interaction between LM and GNN. Compared with prior works, we retain the individual structure of both modalities, consider fine-grained interaction between any token in question and any entity in KG through dense bidirectional attention, and perform multi-step joint reasoning by stacking several interaction layers. Furthermore, we gradually prune the KG size in each stacked model layer under the guidance of attention weights generated in the interactions, making the reasoning path transparent and interpretable.

\section{Methodology}
In this section, we introduce the task definition (\cref{def}) and our JointLK model. The model framework is shown in Figure \ref{model}. JointLK takes the query and the retrieved knowledge subgraph as input, and outputs a real value as the correctness score of the answer. The model is mainly composed of four parts: query encoder, GNN layer, joint reasoning module and dynamic pruning module, of which the latter three form a stack of N identical layers. We use a pre-trained language model to learn the query representation (\cref{query}), and use the GNN layer to learn the graph representation (\cref{gnn}). The Joint Reasoning Module receives these two modalities' representations and then apply dense bidirectional attention to make information fusion and representation update for each token and node (\cref{joint}). The LM-to-KG attention weights generated in reasoning represents the global importance of each node in the graph, so the dynamic pruning module prunes the graph layer by layer according to this weights and finally retains the most relevant nodes (\cref{prune}). After N layers of iteration, the query representation and the trimmed graph representation are used to predict the answer (\cref{predict}).

\subsection{Task Definition} 
\label{def}
The CSQA task in this paper is a multiple-choice problem with some answer choices. Given a commonsense question $q$ and a set of answer choices $\{a_1, a_2, ..., a_n\}$, our task is to measure the plausibility score between $q$ and each answer choice $a$ then select the answer with the highest plausibility score. In general, questions do not contain any reference to answer choices, so the external knowledge graph provides the necessary background knowledge. We extract from the external KG a subgraph $g=(V, R) $ with the guidance of question and choice. Here V is a subset of entity nodes retrieved from the external KG. $E\subseteq{V \times R \times V}$ is the set of edges that connect nodes in $V$, where $R$ is a set of relations types. We describe the detailed extraction process in Appendix~\ref{retrieval}.

\subsection{Query Encoder} 
\label{query}
We follow baselines to use pre-trained language models to encode the query $\{w_{i}\}_{i=1}^{M}$ (question and choice) into a sequence of vectors $\{q_{i}^{0}\}_{i=1}^{M}$:
\begin{equation}
\{\widetilde{q}_{1}^0,...,\widetilde{q}_{M}^0\} = \mathbf{Enc_{LM}}\left(\{w_{1},...,w_{M}\}\right)
\end{equation}
Here $\{\widetilde{q}_{i}^0\}_{i=1}^{M}\in{\mathbb{R}^{T}}$ is the last hidden layer vector of each token in the query. Then we feed the representation of tokens into a non-linear layer so that the text representation space is aligned to the entity representation space:
\begin{equation}
q_i^0=\sigma\left(f_s\left(\widetilde{q}_i^0 \right)\right)
\end{equation}
where $f_s:\mathbb{R}^T\rightarrow\mathbb{R}^D$ is a linear transformation, and $\sigma$ is the activation function. The representations of tokens $\mathbf{Q}^0=\{{q}_{i}^0\}_{i=1}^{M}\in{\mathbb{R}^{D}}$ will be provided to the joint reasoning module for further interaction with the graph entities representations.

\subsection{GNN Layer} 
\label{gnn}
After obtaining token representations by the query encoder, we further model the subgraph to obtain entity representations. First, We use the BERT model with average pooling to get the initial representation for each entity $\mathbf{X^0}=\{x_i^{0}\}_{i=0}^{|V|}\in{\mathbb{R}^{D}}$. Then, we apply GNN Layer to update node representation through iterative message passing between neighbors on the graph, while GNN is built on the RGAT \cite{wang-etal-2020-relational} and is a simplification of \citet{yasunaga-etal-2021-qa}. For brevity, we formulate the entire computation in one layer as:
\begin{equation}
\{\widetilde{x}_1^{l}, ..., \widetilde{x}_{|V|}^{l}\} = \mathbf{GNN}(\{x_1^{l-1}, ..., x_{|V|}^{l-1} \})
\end{equation}
The output representation $x_i^{l}$ is computed by
\begin{align}
\hat{\alpha}_{ji} &= (x_i^{l-1} W_q)(x_j^{l-1} W_k + r_{ji})^{T}, \\
{\alpha}_{ji} &= \mathbf{softmax}(\hat{\alpha}_{ji} / {\sqrt D}), \\
\hat{x}_{i}^{l-1} &= \sum_{j\in N_i\cup\{i\}}{{\alpha}_{ji} (x_j^{l-1} W_v + r_{ji})}, \\
\widetilde{x}_i^{l} &= \mathbf{LayerNorm}(x_i^{l-1} + \hat{x}_{i}^{l-1} W_o)
\end{align}
where matrices $W_q,W_k,W_v,W_o\in\mathbb{R}^{D\times D}$ are trainable parameters, $N_i$ is the neighbor of node $i$. $r_{ji} = \psi(e_{ji},u_j,u_i)$ is the relation feature vector, where $e_{ji}$ is a one-hot vector denoting the relation type of the edge $(j, i)$ and $u_j,u_i$ are one-hot vectors denoting the node types of $j$ and $i$. The following joint reasoning module will further fuse $\widetilde{x}_{i}^{l}$ and $q_{i}^{l-1}$ to obtain their updated representations.

\subsection{Joint Reasoning Module}
\label{joint}
To reduce the gap of query and knowledge graph features, we fuse them in the joint reasoning module by the dense bidirectional attention mechanism that connects two encoding layers of query and knowledge graph and captures the fine-grained interplay between them. 

The module takes the query and KG representations $\mathbf{Q}$ and $\mathbf{X}$ as inputs and then outputs their updated versions. We denote the inputs to the joint reasoning module in the l-st fusion layer by $\mathbf{Q}^{l-1}=\{q_i^{l-1}\}_{i=1}^{M}$ and $\mathbf{\widetilde{X}}^l=\{\widetilde{x}_i^l\}_{i=1}^{|V|}$. Given $q_i^{l-1}$ and $\widetilde{x}_i^l$, an affinity matrix is first constructed via:
\begin{equation}
S_{ij}^{l}=W_S^T[q_i^{l-1};\widetilde{x}_j^{l};q_i^{l-1}\circ \widetilde{x}_j^{l}]
\end{equation}
where $W_S^T$ is a learnable weight matrix, $\circ$ is elementwise multiplication, [;] is vector concatenation across row. We normalize $S_{ij}^{l}$ in row-wise to derive \textbf{KG-to-LM attention} maps on query tokens conditioned by each entity in KG as
\begin{equation}
S_{q_{i}}^{l} = \mathbf{softmax}\left(S_{ij}\right)
\end{equation}
and also normalize $S_{ij}^{l}$ in column-wise to derive \textbf{LM-to-KG attention} maps on entities conditioned by each query token as
\begin{equation}
\label{lm-to-kg}
S_{x_{j}}^{l} = \mathbf{softmax}\left(S_{ij}^{T}\right)
\end{equation}
The attended representations are computed as follows:
\begin{equation}
\hat{q}_{ij} = q_{i}^{l-1}\otimes S_{q_{i}}^{l},\\
\hat{x}_{ij} = \widetilde{x}_{j}^{l}\otimes S_{x_{j}}^{l}
\end{equation}
where $\otimes$ represents matrix multiplication. The attended features are fused with the original features of the other modality by concatenation and then compressed to low-dimensional space by:
\begin{align}
q_i^{l}&=W_Q [q_i^{l-1} ; \hat{x}_{ij}; q_i^{l-1} \circ \hat{x}_{ij}; q_i^{l-1}\circ \hat{q}_{ij}], \\
\bar{x}_j^l&=W_X [\widetilde{x}_j^l ; \hat{q}_{ij}; \widetilde{x}_j^l \circ \hat{q}_{ij} ; \widetilde{x}_j^l \circ \hat{x}_{ij}]
\end{align}
where $W_Q,W_X$ are learnable weights. 
Then the updated query representation $\mathbf{Q}^{l}=\{q_i^{l}\}_{i=1}^{M}$ will be input to the next $l$-th stacked JointLK layer of to continue participating in joint reasoning, and the updated KG representation
$\mathbf{\bar{X}}^l=\{\bar{x}_i^l\}_{i=1}^{|V|}$ will be input to the next module of the current JointLK layer for pruning.

\subsection{Dynamic Pruning Module}
\label{prune}
In Equation \ref{lm-to-kg}, the LM-to-KG attention value implies the importance of different nodes in the subgraph for question answering. Inspired by SAGPool \cite{pmlr-v97-lee19c}, under the guidance of query, we retain relevant nodes and cut out irrelevant nodes according to the LM-to-KG attention. Then, We define a hyperparameter, the Retention ratio $K\in(0,1]$, which determines the number of nodes to be retained. We choose the top $\left\lceil K\cdot |V|\right\rceil$ nodes according to the value of LM-to-KG attention:
\begin{gather}
idx = \mathbf{top-rank}\left(Z,\left\lceil K\cdot |V|\right\rceil\right),\\ Z_{mask} = Z_{idx}
\end{gather}
where top-rank is a function that returns the index of top $\left\lceil K \cdot |V| \right \rceil$ value, $\cdot_{idx}$ is an indexing operation, and $Z_{ mask}$ is corresponding attention mask. Next, the subgraph is formed by pooling out the less essential entity nodes as:
\begin{equation}
\begin{split}
\mathbf{X}^{l} &= \mathbf{\bar{X}}_{idx,:}^{l}\odot Z_{mask},\\ 
\mathbf{A}^{l} &= \mathbf{\bar{A}}_{idx,idx}^{l}
\end{split}
\end{equation}
where $\mathbf{\bar{X}}_{idx,:}^{l}$ is the row-wise indexed representation matrix of $\mathbf{\bar{X}}^{l}$, $\odot$ is the broadcasted elementwise product, and  $\mathbf{\bar{A}}_{idx,idx}^{l}$ is the row-wise and col-wise of indexed adjacency matrix. $\mathbf{X}^{l}=(x_1^{l},x_2^{l},\ldots,x_{ \left\lceil k |V|\right\rceil}^{l})$, $\mathbf{A}^{l}$ and $\left\lceil K\cdot |V|\right\rceil$ are the representation matrix, the adjacency matrix and the number of graph nodes in the next JointLK layer.

\subsection{Answer Prediction}
\label{predict}
After N layers of iteration, we finally obtain the query representation $\mathbf{Q}^{N}$ that fuses knowledge information and the graph representation $\mathbf{X}^{N}$ that fuses question information. 
We compute the score of $a$ being the correct answer as:
\begin{equation}
p=\left(a|q\right) = \mathbf{MLP}\left([{s}; {g}]\right)
\end{equation}
where $s$ is the mean pooling of $\mathbf{Q}^{N}$, and $g$ is the attention-based pooling of $\mathbf{X}^{N}$. 
We get the final probability by normalize all question-choice pairs with softmax.

\section{Experimental Setup}
\subsection{Datasets}
We evaluate our model on two typical commonsense question answering datasets CommonsenseQA \cite{talmor-etal-2019-commonsenseqa} and OpenBookQA \cite{mihaylov-etal-2018-suit}. \textbf{CommonsenseQA} is a 5-way multiple-choice question answering dataset that requires commonsense for reasoning and contains 12,102 questions. We experiment and report the accuracy on the in-house dev (IHdev) and test (IHtest) splits used by \citet{lin-etal-2019-kagnet}, and report the accuracy of our final system on the official test set.
\textbf{OpenBookQA} is a 4-way multiple choice question answering dataset that requires reasoning with elementary science knowledge. It contains 5,957 questions along with an open book of scientific facts. We use the official data split.

\subsection{Implementation Details}
Following previous work \cite{yasunaga-etal-2021-qa}, we use ConceptNet \cite{speer2017conceptnet}, a commonsense knowledge graph, as our structured knowledge source for both of the above tasks. Given each query, we follow the preprocessing steps described in \citet{feng-etal-2020-scalable} to retrieve the subgraph from ConceptNet, and the max hop size is 3 (see Appendix \ref{retrieval} for the detail). We use cross-entropy loss and RAdam optimizer \cite{Liu2020On}. In training, we set the maximum input sequence length to text encoders to 100, batch size to 128, and perform early stopping. We set the dimension (D = 200) and number of layers (N = 5) of our GNN module, with dropout rate 0.2 applied to each layer \cite{JMLR:v15:srivastava14a}. We use separate learning rates for the LM encoder and the graph encoder. We choose the LM encoder learning rate from$\{1\times{10}^{-5},\ 2\times{10}^{-5},\ 3\times{10}^{-5}\}$, and choose the graph encoder learning rate from$\{1\times{10}^{-3},\ 2\times{10}^{-3}\}$. Each model is trained using one GPU (Tesla\_v100-sxm2-16gb), which takes 20 hours on average.

\subsection{Compared Method}
Although text corpus can provide complementary knowledge except for knowledge graphs, our model focuses on improving the use of KG and the joint reasoning between LM and KG, so we choose LM and LM+KG as the comparison methods.

To investigate the role of KGs, we compare with the benchmark model RoBERTa-large \cite{liu2019roberta} for CommonsenseQA, and compare with RoBERTa-large and AristoRoBERTa \cite{clark2020f} for OpenBookQA. For LM+KG methods, they share a similar high-level framework with our methods, that is, LM is used as a text encoder, GNN or RN is used as a KG encoder, but the way of using knowledge or reasoning is different: (1) Relationship network (RN) \cite{NIPS2017_e6acf4b0}, (2) RGCN \cite{schlichtkrull2018modeling}, (3) GconAttn \cite{wang2019improving}, (4)KagNet \cite{lin-etal-2019-kagnet} and (5)MHGRN \cite{feng-etal-2020-scalable}, (6) QA-GNN \cite{yasunaga-etal-2021-qa}. (1), (2) and (3) are the relational perception GNNs for KGs, and (4), (5) and (6) are further model paths in KGs. To be fair, we use the same LM for all comparison methods. 

\begin{table}[t]
\centering
\resizebox{1\columnwidth}{!}
{
\begin{tabular}{lcc}
\toprule   \textbf{Methods} & \textbf{IHdev-Acc.(\%)} & \textbf{IHtest-Acc.(\%)} \\
\toprule    RoBERTa-large(w/o KG) & 73.07 ($\pm$0.45) & 68.69 ($\pm$0.56) \\
\hline
+ RGCN & 72.69 ($\pm$0.19) & 68.41 ($\pm$0.66)\\
+ GconAttn & 71.61 ($\pm$0.39) & 68.59 ($\pm$0.96)\\
+ KagNet & 73.47 ($\pm$0.22) & 69.01 ($\pm$0.76)\\
+ RN & 74.57 ($\pm$0.91) & 69.08 ($\pm$0.21)\\
+ MHGRN & 74.45 ($\pm$0.10) & 71.11 ($\pm$0.81)\\
+ QA-GNN & 76.54 ($\pm$0.21) & 73.41 ($\pm$0.92)\\
\midrule
+ JointLK (Ours) & \textbf{77.88} ($\pm$0.25) & \textbf{74.43} ($\pm$0.83)\\
\bottomrule
\end{tabular}
}
\caption{Performance comparison on CommonsenseQA in-house split. We follow the data division method of \citet{lin-etal-2019-kagnet} and report the in-house Dev (IHdev) and Test (IHtest) accuracy(mean and standard deviation of four runs).}
\label{table1}
\end{table}

\begin{table}[t]
\centering
\resizebox{1\columnwidth}{!}{
\begin{tabular}{lc}
\toprule   \textbf{Methods} & \textbf{Test} \\
\toprule    
RoBERTa \cite{liu2019roberta} & 72.1 \\
\hline
Albert \cite{Lan2020ALBERT:} (ensemble) & 76.5\\
\hline
RoBERTa + FreeLB \cite{Zhu2020FreeLB:} (ensemble) & 73.1\\
RoBERTa + HyKAS \cite{ma-etal-2019-towards} & 73.2\\
RoBERTa + KE (ensemble) & 73.3\\
RoBERTa + KEDGN (ensemble) & 74.4\\
XLNet + GraphReason \cite{lv2020graph} & 75.3\\
RoBERTa + MHGRN \cite{feng-etal-2020-scalable} & 75.4\\
Albert + PG \cite{wang-etal-2020-connecting} & 75.6\\
RoBERTa + QA-GNN \cite{yasunaga-etal-2021-qa} & 76.1\\
\midrule
RoBERTa + JointLK (Ours) & \textbf{76.6}\\
\bottomrule
\end{tabular}
}
\caption{Performance comparison on the CommonsenseQA official leaderboard. Our model has achieved state-of-the-art under the setting of RoBERTa-large.}
\label{table2}
\end{table}

%OpenBookQA
\begin{table}[t]
\centering
\resizebox{1\columnwidth}{!}
{
\begin{tabular}{lcc}
\toprule   \textbf{Methods} & \textbf{RoBERTa-large} & \textbf{AristoRoBERTa} \\
\toprule    Fine-tuned LMs (w/o KG) & 64.80 ($\pm$2.37) & 78.40 ($\pm$1.64) \\
\hline
+ RGCN & 62.45 ($\pm$1.57) & 74.60 ($\pm$2.53)\\
+ GconAttn & 64.75 ($\pm$1.48) & 71.80 ($\pm$1.21)\\
+ RN & 65.20 ($\pm$1.18) & 75.35 ($\pm$1.39)\\
+ MHGRN & 66.85 ($\pm$1.19) & 80.6\\
+ QA-GNN & 67.80 ($\pm$2.75) & 82.77 ($\pm$1.56)\\
\midrule
+ JointLK (Ours) & \textbf{70.34} ($\pm$0.75) & \textbf{84.92} ($\pm$1.07)\\
\bottomrule
\end{tabular}
}
\caption{Test accuracy on OpenBookQA. Methods with
AristoRoBERTa use the textual evidence by \citet{clark2020f} as an additional input to the QA context.}
\label{obqa1}
\end{table}

%OpenBookQA,test
\begin{table}[t]
\centering
\resizebox{1\columnwidth}{!}{
\begin{tabular}{lc}
\toprule   \textbf{Methods} & \textbf{Test} \\
\toprule    
Careful Selection \cite{banerjee-etal-2019-careful} & 72.0 \\
AristoRoBERTa & 77.8\\
KF + SIR \cite{banerjee2020knowledge} & 80.0\\
AristoRoBERTa + PG \cite{wang-etal-2020-connecting} & 80.2\\
AristoRoBERTa + MHGRN \cite{feng-etal-2020-scalable} & 80.6\\
ALBERT + KB & 81.0\\
AristoRoBERTa + QA-GNN \cite{yasunaga-etal-2021-qa} & 82.8\\
T5* \cite{JMLR:v21:20-074} & 83.2\\
UnifiedQA(11B)* \cite{khashabi-etal-2020-unifiedqa} & 87.2\\
\midrule
AristoRoBERTa + JointLK (Ours) & \textbf{85.6}\\
\bottomrule
\end{tabular}
}
\caption{Test accuracy on OpenBookQA leaderboard. All listed methods use the provided science facts as an additional input to the language context. The previous top 2 systems, UnifiedQA (11B params) and T5 (3B params) are 30x and 8x larger than our model.}
\label{obqa2}
\end{table}

\begin{table}[t]
\centering
\resizebox{0.8\columnwidth}{!}{
\begin{tabular}{lc}
\toprule   \textbf{Methods} & \textbf{IHdev-Acc. (\%)} \\
\hline
JointLK ($N$=5) & \textbf{77.88}\\
- Dynamic Pruning Module	& 77.38\\
- Joint Reasoning Module	& 76.61\\
\bottomrule
\end{tabular}
}
\caption{Ablation study on model components using RoBERTa-large as the text encoder. We report the IHdev accuracy on CommonsenseQA.}
\label{table3}
\end{table}

\begin{figure}[t]
\centering    %居中
\subfigure[] %第一张子图
{
	\begin{minipage}{3.61cm}
% 	\centering          %子图居中
	\includegraphics[scale=0.54]{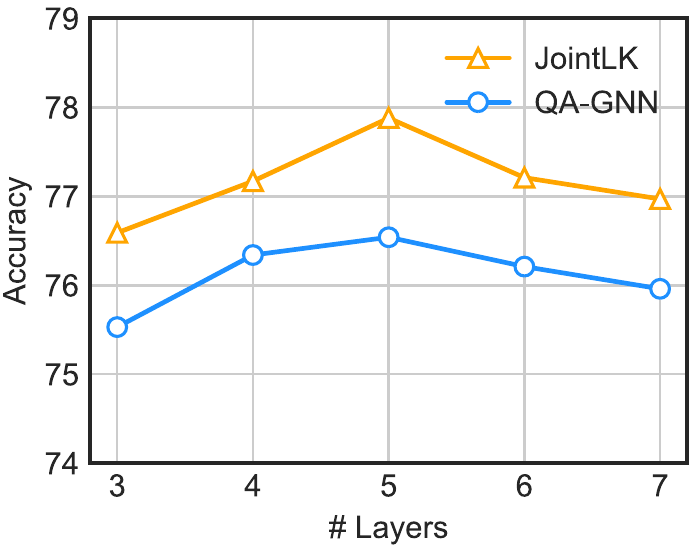}   %以pic.jpg的0.5倍大小输出
	\end{minipage}
}
\subfigure[] %第二张子图
{
	\begin{minipage}{3.61cm}
% 	\centering      %子图居中
	\includegraphics[scale=0.52]{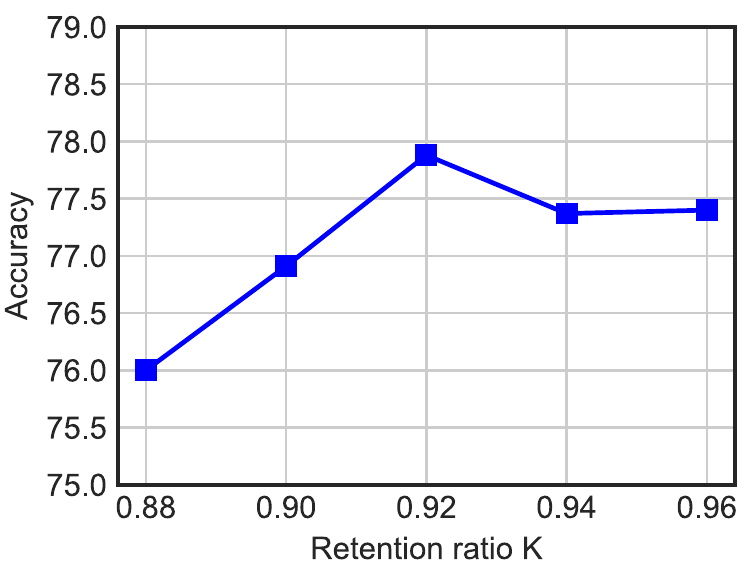}   %以pic.jpg的0.5倍大小输出
	\end{minipage}
}
\caption{Ablation study on stacked of JointLK layers (a) and the retention ratio in pruning (b).} %  %大图名称
\label{NK}  %图片引用标记
\end{figure}

\section{Results and Analysis}
\subsection{Main Results}
The results on CommonsenseQA in-house split dataset and official test dataset are shown in Table~\ref{table1} and Table \ref{table2}. The results on OpenBookQA test dataset and leaderboard are shown in Table \ref{obqa1} and Table \ref{obqa2}. We can observe that JointLK performs best among all fine-tuned LMs and existing LM+KG models. On CommonsenseQA, our model’s test performance improves by 5.74\% over fine-tuned LMs and 1.02\% over the prior best LM+KG model, QA-GNN. On OpenbookQA, our model’s test performance improves by 6.52\% over fine-tuned AristoRoBERTa, and 2.15\% over QA-GNN. Additionally, we also submit our best model to the leaderboards, and our JointLK (with the text encoder being RoBERTa-large) ranks first among comparable approaches. Compared with the previous best model MHGRN and QA-GNN, the boost over them suggests the effectiveness of our proposed joint reasoning between LM and KG and the dynamic pruning mechanism. 

In particular, we do not compare with the higher ranking models on the leaderboard, such as unified QA \cite{khashabi-etal-2020-unifiedqa}, Albert + DESC-KCR \cite{xu-etal-2021-fusing}, because they either use a stronger text encoder or use additional data resources, while our model focuses on improving the joint reasoning between LM and KG.

\begin{table}[t]
\centering
\resizebox{1\columnwidth}{!}
{
\begin{tabular}{p{1.8cm} p{1.8cm}<{\centering} p{2.1cm}<{\centering} p{2.2cm}<{\centering} p{2.2cm}<{\centering}}
\toprule   \textbf{Methods} & \textbf{IHdev-Acc (Overall)} & \textbf{IHdev-Acc (Questions w/ negation)} & \textbf{IHdev-Acc (Questions w/ $\leq$7 entities)} & \textbf{IHdev-Acc (Questions w/ >7 entities)}\\
\toprule    \textit{Number} & \textit{1221} & \textit{133} & \textit{723} & \textit{498} \\
\midrule
QA-GNN & 76.99 & 72.18 & 76.63 & 77.51\\
\midrule
JointLK(\textbf{Ours}) & 78.38 & \textbf{75.18  ($\uparrow$3.00)} & \textbf{77.59  ($\uparrow$0.96)} & \textbf{79.52  ($\uparrow$2.01)}\\
\bottomrule
\end{tabular}
}
\caption{Performance on questions with negative words and fewer/more entities. The questions are retrieved from the CommonsenseQA IHdev set.}
\label{question}
\end{table}

\subsection{Ablation Studies}
We further conduct in-depth analyses to investigate the effectiveness of different components in our model. We show the accuracy of JointLK on the CommonsenseQA IHdev set.

\noindent\textbf{Impact of JointLK components}
We assess the impact of the joint reasoning module (\cref{joint}) and the dynamic pruning module (\cref{prune}), shown in Table~\ref{table3}. Disabling the dynamic pruning module results in 0.5\% drop in performance, showing that some nodes in subgraph are not conducive to reasoning. Especially, when we disable the joint reasoning module, the corresponding dynamic pruning module will also be removed, because the latter depends on the attention value in the former. Then the results have a significant drop: $77.88\% \rightarrow 76.61\%$, suggesting that the joint reasoning between LM and KG is critical.

\noindent\textbf{Impact of stacked of JointLK Layers}
We investigate the impact of the number of JointLK layers (shown in Figure \ref{NK} (a)). The increase of layers continues to bring benefits until layers $N= 5$. However, performance begins to drop when $N > 5$. As the number of layers increases, the model changes from underfitting to overfitting.

\noindent\textbf{Impact of the Retention Ratio in Pruning}
The retention ratio $K$ is a hyperparameter of the dynamic pruning module. Since it is recursively pruning in each stacked layer of JointLK, the percentage of graph nodes that the model ultimately retains is also related to the number of layers of JointLK, that is, $K^N$, where $N=5$. Experiments show that if the retention ratio is too high, there may be almost no pruning effect (for example, K=0.98, 90\% of the nodes are retained in the last layer); otherwise, useful nodes may be deleted. As shown in Figure \ref{NK} (b), when the number of JointLK layers $N=5$, $K=0.92$ (about 66\% of the original nodes remain in the last layer) works the best on the CommonsenseQA dev set.

\subsection{Quantitative Analysis}
Considering the overall performance improvement of our model on these two datasets, we analyze whether the improvement is reflected in questions that require more complex reasoning, such as questions with negation and complex questions with more entities. We compare our model with the prior best LM+KG model, QA-GNN in Table \ref{question}.

\noindent \textbf{Questions with negation}
Large LMs do well due to memorizing subject and filler co-occurrences but are easily distracted by elements like negation \cite{zagoury2021s}. To investigate the reasoning ability of the model on negation, we retrieved 133 questions with negation terms (e.g., no, not, nothing, never, unlikely, don't, doesn't, didn't, can't, couldn't) from the CommonsenseQA IHdev set.  JointLK exhibits a big boost ($\uparrow$3.00\%) over QA-GNN, suggesting its strength in negation reasoning. The fine-grained joint inference of LM and GNN allows the model to pay attention to the semantic nuances of language expressions.

\noindent \textbf{Questions with fewer/more entities}
When the question contains many entities, the size and noise of the retrieved KG may limit the model's performance because the model needs to understand the complex relationship between entities. According to statistics (see Appendix \ref{retrieval}), questions contain an average of 7 entities, so we divide the question into two categories: containing fewer entities ($\leq$7) and more entities($>$7). Compared with QA-GNN, JointLK has a bigger boost on questions with more entities ($\uparrow$2.01\%) than those with fewer entities ($\uparrow$0.96\%), suggesting that our model can reduce the reasoning difficulty of complex questions because it can remove irrelevant nodes in reasoning.

\begin{figure*}[t]
\centering
\includegraphics[width=1\textwidth]{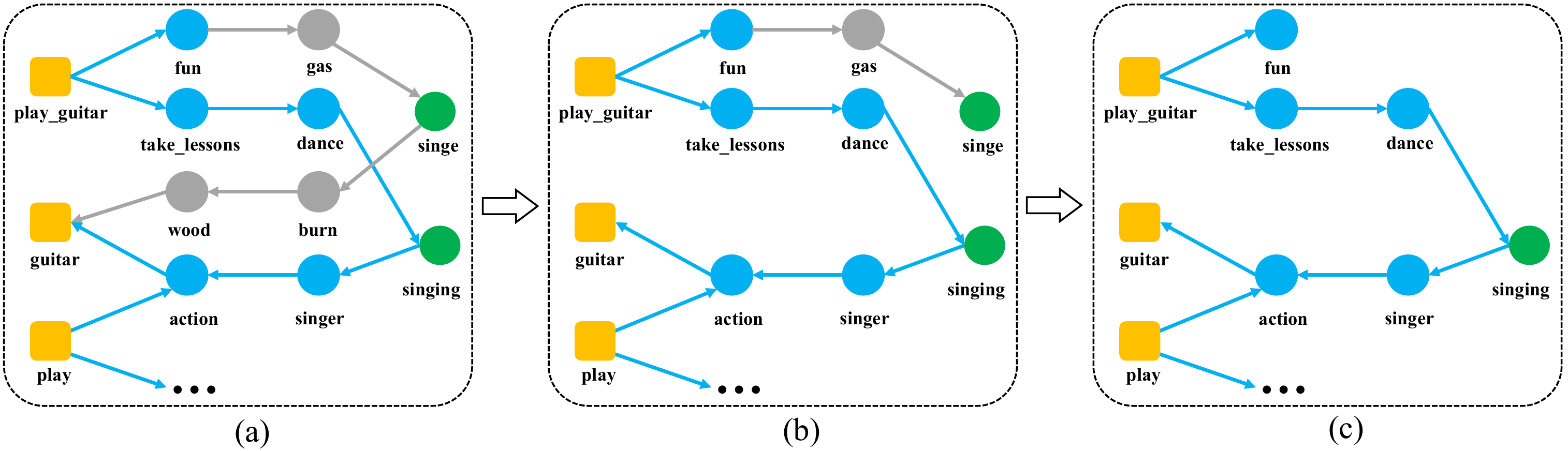} % Reduce the figure size so that it is slightly narrower than the column.
\caption{Case study of our model reasoning and pruning process. The question and answer choices corresponding to this case are: "What do people typically do while playing guitar? A.cry B. hear sounds C. \textbf{singing} D. arthritis  E. making music".}
\label{case}
\end{figure*}

\subsection{Interpretability: A Case Study}
We aim to interpret JointLK's reasoning process by analyzing the pruning of the knowledge subgraph. Figure \ref{case} shows an example from CommonsenseQA where our model correctly answers the question and finally retains reasonable reasoning paths by pruning the subgraph. The flow from (a) to (b) to (c) represents the recursive pruning of the subgraph according to the LM-to-KG attention weight at each GNN update layer. From (a) to (b), although the nodes \textit{wood} and \textit{burn} bridge the reasoning gap between question entity and answer entity, their semantics are very different from the question. From (b) to (c), ``$play\_guitar\stackrel{usedfor}{\longrightarrow}fun$" and ``$fun\stackrel{relatedto}{\longrightarrow}gas\stackrel{relatedto}{\longrightarrow}singe$" are both reasonable, but the former is related to the semantics of the question, and the latter is not. Two paths are reserved in (c), ``$play\_guitar\stackrel{hassubevent}{\longrightarrow}take\_lessons\stackrel{hassubevent}{\longrightarrow}dance\stackrel{relatedto}{\longrightarrow}singing"$ and ``$play\_guitar\stackrel{relatedto}{\longrightarrow}action\stackrel{relatedto}{\longrightarrow}singer\stackrel{relatedto}{\longrightarrow}singing$". These two paths describe two possible scenarios that support answering the question.

\subsection{Error Analysis}
In order to understand why our model fails in some cases, we randomly select 100 error cases and group them into several categories. There are three main types of errors, and we show some examples in the Appendix \ref{error}.

\noindent\textbf{Miss important evidence (39/100)}
Although we can retrieve many nodes related to questions and choices from ConceptNet, due to the incompleteness of the knowledge graph, there may be missing essential evidence nodes in the reasoning paths to answer the question. For example, although \textit{``eating\_dinner"} will cause \textit{``sleepiness"} or \textit{``indigestion"}, knowledge such as ``lactose intolerance causes indigestion" is essential to answer the question (Wikipedia: \textit{Lactose intolerance is a common condition caused by a decreased ability to digest lactose, a sugar found in dairy products.}). However, ConceptNet does not cover such knowledge or not is retrieved.

\noindent\textbf{Indistinguishable knowledge (25/100)}
Several choices of the question may be correct, difficult to distinguish, and which one is correct may vary from person to person. For example, \textit{``human"} and \textit{``cat"} may be at location \textit{``bed"} or \textit{``comfortable chair"}, and the knowledge provided by ConceptNet is also the same. The model may choose bed because the bed appears more frequently in the pre-trained corpus.

\noindent\textbf{Incomprehensible questions (23/100)}
This type of error often occurs when the question is particularly long, involving various events and changes in the characters' emotions. The model is difficult to understand the scene described by the question. Some questions may require reasoning based on events, but the knowledge in ConceptNet is more based on entities and attributes.

The above three types of errors show that selecting complete, accurate, and context-sensitive knowledge is vital for more effective KG-augmented models.

\section{Conclusion}
In this work, we propose JointLK and provide a set of experiments to prove that (i) LM and KG interactive fusion can reduce the semantic gap between the two information modalities and make better use of KG for joint reasoning with LM. (ii) Dynamic pruning module can recursively delete irrelevant subgraph nodes at each layer of JointLK to provide fine appropriate evidence. Our results on CommonsenseQA and OpenBookQA demonstrate the superiority of JointLK over other methods using external knowledge and the strong performance in performing complex reasoning. In addition, our research results can be broadly extended to other tasks that require KGs as additional background knowledge to augment LMs, such as entity linking, KG completion and the recommendation system.

\section*{Acknowledgements}
We would like to thank the anonymous reviewers for their helpful comments. This work was supported by the Key Development Program of the Ministry of Science and Technology (No.2019YFF0303003), the National Natural Science Foundation of China (No.61976068) and "Hundreds, Millions" Engineering Science and Technology Major Special Project of Heilongjiang Province (No.2020ZX14A02).

\section*{Ethical Impact}
This paper proposes a general approach to fuse language models and external knowledge graphs for commonsense reasoning. We worked within the purview of acceptable privacy practices and strictly followed the data usage policy. In all the experiments, we use public datasets and consist of their intended use. We neither introduce any social/ethical bias to the model nor amplify any bias in the data, so we do not foresee any direct social consequences or ethical issues.

\bibliography{acl}
\bibliographystyle{acl}

\appendix

\section{Extracting subgraph from External KG}
\label{retrieval}
We choose ConceptNet as the external knowledge base, and we follow the process of \citet{feng-etal-2020-scalable} and \citet{yasunaga-etal-2021-qa} to retrieve the knowledge subgraph.

Given the question and choice, we identify the concepts that appear in ConceptNet in question and choice, respectively, and get the initial node set $V_q$ and $V_a$, which form the initial node set $V_{q,a}$. For example, in the question ``\textit{What do people typically do while playing guitar?}" and choice ``\textit{singing}", $V_q$ = \{guitar, people, play, play\_guitar, playing, playing\_guitar, typically\}, $V_a$ = \{singe, singing\}. Then, in order to extract the subgraph related to question and choice, we add the bridge entities on the 1 and 2 hop paths between any pair of entities in $V_{q,a}$, thus obtaining the retrieved entity set $V$.

There may be many nodes in $V$, especially long questions contain many concepts. We follow the preprocessing method of \citet{yasunaga-etal-2021-qa}, connect the nodes with question + choice, and calculate the relevant scores of the nodes through a pre-trained LM. We only retain the top 200 scoring nodes (It is worth noting that this is the preprocessing of the retrieval process, which is different from the dynamic pruning in section \ref{prune}. The former is to score only one node and separate from the whole subgraph where the node is located, while the latter is recursive pruning in the updating process of the modeling subgraph).

Finally, we get the relation set $R$ by merging the relation types in ConceptNet and adding reverse relation. We retrieve all the edges in $R$ of any two nodes in $V$. In addition, we add question as a node $q$ to $V$, and add the bidirectional edges of $q$ to $V_q$ and $q$ to $V_a$. The relation types are shown in Table~\ref{relation}, and the statistics of the retrieved nodes are shown in Table \ref{nodes}.

\begin{table}[p]
\centering
\resizebox{0.8\columnwidth}{!}{
\begin{tabular}{c|c}
\toprule   Relation & Merged Relation \\
\hline
AtLocation & \multirow{2}{*}{AtLocation}\\
LocatedNear &   \\
\hline

Causes	& \multirow{3}{*}{Causes}\\
CausesDesire &   \\
*MotivatedByGoal &   \\
\hline

Antonym   &   \multirow{2}{*}{Antonym}\\
DistinctFrom &  \\
\hline

HasSubevent  &   \multirow{6}{*}{HasSubevent}\\
HasFirstSubevent  &   \\
HasLastSubevent  &   \\
HasPrerequisite  &   \\
Entails  &   \\
MannerOf  &   \\
\hline

IsA  &   \multirow{3}{*}{IsA}\\
InstanceOf  &   \\
DefinedAs  &   \\
\hline

PartOf   &   \multirow{2}{*}{PartOf}\\
*HasA   &   \\
\hline

RelatedTo   &   \multirow{3}{*}{RelatedTo}\\
SimilarTo   &   \\
Synonym   &   \\
\hline

CapableOf	& CapableOf\\
\hline
CreatedBy   & CreatedBy\\
\hline
Desires &   Desires\\
\hline
UsedFor &   UsedFor\\
\hline
HasContext  &  HasContext \\
\hline
HasProperty &   HasProperty\\
\hline
MadeOf  &   MadeOf\\
\hline
NotCapableOf    &   NotCapableOf\\
\hline
NotDesires  &  NotDesires \\
\hline
ReceivesAction  &  ReceivesAction \\
\hline
$q\rightarrow V_{q}$ & $q\rightarrow V_{q}$ \\
\hline
$q\rightarrow V_{a}$ & $q\rightarrow V_{a}$ \\
\bottomrule
\end{tabular}
}
\caption{Relation types after preprocessing. *RelationX indicates the reverse relation of RelationX. There are 19 kinds of merged relations. We consider the reverse edge of each relation during training and testing, so there are 38 relation types in total.}
\label{relation}
\end{table}

\begin{table*}[p]
\centering
\resizebox{1.6\columnwidth}{!}{
\begin{tabular}{l|c|c|c|c}
\toprule   Datesets & Split & Average $|V_q|$ & Average $|V_a|$ & Average $|V|$ \\
\hline
\multirow{3}{*}{CommomsenseQA}
 & Train set & 7.43 & 2.07 & 107.96 \\
 & Dev set   & 7.20 & 2.05 & 106.55 \\
 & Test set  & 7.38 & 2.05 & 106.22 \\
\hline
\multirow{3}{*}{OpenBookQA}
 & Train set & 6.59 & 2.85 & 100.14 \\
 & Dev set   & 6.48 & 3.41 & 108.15 \\
 & Test set  & 6.42 & 3.08 & 101.60 \\

\bottomrule
\end{tabular}
}
\caption{Statistics on the number of retrieved subgraph nodes corresponding to each piece of data. $V_q$ is the set of entities included in a question. $V_a$ is the set of entities included in a choice. $V$ contains $V_q$, $V_a$, and any bridging entity with no more than two hops between any pair of entities in $V_q$ and $V_a$.}
\label{nodes}
\end{table*}

\section{Node Initialization}
\label{node initial}
For each entity in the subgraph, we need to obtain its feature representation. Following \cite{feng-etal-2020-scalable}, we first use the template to convert the knowledge triples in ConceptNet into sentences, and feed them into BERT-Large, obtaining a sequence of tokens embeddings from the last layer. For each entity, we perform mean pooling over the tokens of the entity's occurrences across all the sentences to form the initial embeddings $x_i^{0}$.

% \section{The performance of JointLK with other LMs}
% \label{plm}

\section{Error Types and Examples}
\label{error}
In Table \ref{error table}, we present examples for each error type in the Commonsense IHdev set. Because the average number of subgraph nodes corresponding to each case is about 100, we cannot list them all. Only some important nodes are shown here.

\begin{table*}[p]
\centering
\resizebox{2\columnwidth}{!}{
\begin{tabular}{l|lp{10cm}}
\toprule   
\textbf{Error type} & \textbf{Example} &  \\
\hline
\multirow{4}{2.6cm}{\textbf{Missing important evidence (39/100)}} & Question &  He has lactose intolerant, but was \textbf{eating dinner} made of cheese, what followed for him?\\
 & Answer choices &  digestive$\times |$  feel better $\times |$  \textbf{sleepiness}$\times |$  \textbf{indigestion} $\checkmark |$   illness$\times$\\
 & Subgraph for correct answer &  $eating\_dinner \stackrel{causes}{\longrightarrow} indigestion$, $intolerant\stackrel{relatedto}{\longrightarrow}pain\stackrel{isa}{\longrightarrow}symptom\stackrel{isa}{\longleftarrow}indigestion $, ...\\
 & Subgraph for predicted answer & $lactose \stackrel{relatedto}{\longrightarrow}food \stackrel{hassubevent}{\longrightarrow}eating\_dinner \stackrel{causes}{\longrightarrow}sleepiness $, $ intolerant\stackrel{relatedto}{\longrightarrow}bear\stackrel{relatedto}{\longrightarrow} sleep$, ...\\
\hline
\hline
\multirow{4}{2.6cm}{\textbf{Indistinguishable knowledge (25/100)}} & Question &  Where would a cat snuggle up with their human?\\
 & Answer choices &  floor$\times |$  humane society$\times |$   \textbf{bed}$\times |$   \textbf{comfortable chair}$\checkmark |$  window sill$\times $\\
 & Subgraph for correct answer &  $cat \stackrel{atlocation}{\longrightarrow}chair$, $human \stackrel{atlocation}{\longrightarrow}chair $, ...\\
 & Subgraph for predicted answer & $cat \stackrel{atlocation}{\longrightarrow}bed$, $human \stackrel{atlocation}{\longrightarrow}bed $, ...\\
\hline
\hline
\multirow{4}{2.7cm}{\textbf{Incomprehensible questions (23/100)}} & Question &  The man tried to \textbf{break} the  glass in order to make his escape in time, but he could not. The person in the car, trying to kill him, did what?\\
 & Answer choices &  \textbf{accelerate}$\checkmark |$ putting together$\times |$ \textbf{working}$\times |$ construct$\times |$  train$\times $ \\
 & Subgraph for correct answer &  $escape\stackrel{isa}{\longleftarrow} break\stackrel{antonym}{\longrightarrow}accelerate $, $kill\stackrel{relatedto}{\longrightarrow}attack\stackrel{relatedto}{\longrightarrow}accelerate $, $man\stackrel{relatedto}{\longrightarrow}break\stackrel{relatedto}{\longrightarrow}falling\stackrel{hassubevent}{\longrightarrow}accelerate$, ...\\
 & Subgraph for predicted answer & $break\stackrel{isa}{\longrightarrow}action\stackrel{relatedto}{\longrightarrow}work$, $escape\stackrel{isa}{\longleftarrow}break\stackrel{hassubevent}{\longleftarrow}work $, $kill\stackrel{causes}{\longrightarrow}die\stackrel{hassubevent}{\longleftarrow}work $, ...\\

\bottomrule
\end{tabular}
}
\caption{Several error cases of JointLK model on CommonsenseQA dev dataset. Because there are many nodes in the subgraph, we represent some nodes and relationships in the subgraph in the form of links.}
\label{error table}
\end{table*}

\end{document}